\documentclass[
]{ceurart}

\usepackage{listings}
\usepackage{multirow}
\usepackage{multicol}
\begin{document}

%%
%% Rights management information.
%% CC-BY is default license.
\copyrightyear{2026}
\copyrightclause{Copyright for this paper by its authors.
  Use permitted under Creative Commons License Attribution 4.0
  International (CC BY 4.0).}

%%
%% This command is for the conference information
\conference{OM 2026: The 21st International Workshop on Ontology Matching collocated with the 25th International Semantic Web Conference (ISWC 2026), October 25th, 2026, Bari, Italy}

%% The "title" command
\title{OntoAligner-Ensemble: Voting-Based Fusion across Heterogeneous Ontology Alignment Techniques}
%\title{Ensemble Learning for Ontology Alignment}

\author[1]{Hamed {Babaei Giglou}}[%
orcid=0000-0003-3758-1454,
email=hamed.babaei@tib.eu,
]
\address[1]{TIB Leibniz Information Centre for Science and Technology, Hannover, Germany}

\author[1,2]{Sören Auer}[%
orcid=0000-0002-0698-2864,
email=auer@tib.eu,
]
\address[2]{L3S Research Center, Leibniz University of Hannover, Hannover, Germany}

\author[3]{Peio Popov}[%
email=peio.popov@graphwise.ai
]
\address[3]{Graphwise, Sofia, Bulgaria}

\author[4]{Mahsa Sanaei}[%
orcid=0009-0008-2154-0627,
email=mahsa.san75@gmail.com
]
\address[4]{University of Tabriz, Tabriz, Iran}

\author[1]{Jennifer D'Souza}[%
orcid=0000-0002-6616-9509,
email=jennifer.dsouza@tib.eu,
]

% \thanks{Project website: \url{https://ontoaligner.github.io/}}

%%
%% The abstract is a short summary of the work to be presented in the
%% article.
\begin{abstract}
% Although modern OA frameworks provide unified ecosystems for deploying these heterogeneous aligners, a general mechanism for systematically reconciling their complementary and often conflicting predictions remains an open challenge.
Ontology alignment (OA) has evolved through several methodological paradigms, ranging from lexical and structural aligners to knowledge graph embedding (KGE) models and, more recently, Large Language Model (LLM)-based approaches. Although modern OA frameworks provide unified ecosystems for deploying these heterogeneous aligners, mechanisms for systematically reconciling their complementary and sometimes conflicting predictions remain relatively underexplored. We present OntoAligner-Ensemble, a modular and aligner-agnostic framework that combines candidate correspondences through a configurable two-stage process comprising voting-based fusion strategies (e.g., Weighted Voting, Reciprocal Rank Fusion, Condorcet, and Borda Count) followed by post-fusion selection policies. The framework supports any ontology aligner implemented within OntoAligner that produces candidate correspondences, enabling diverse alignment paradigms to be integrated through a unified decision process. To demonstrate its effectiveness, we instantiate the framework using representative lightweight string-aligner, KGE-based, and Retrieval-Augmented Generation (RAG) aligners powered by both open-weight and API-based LLMs. We evaluate individual aligners and ensemble configurations across eight benchmark tasks from five OAEI tracks spanning biomedical, material science, biodiversity, circular economy, and beyond-equivalence ontology matching. The results show that ensemble fusion consistently improves the balance between precision and recall and frequently outperforms standalone aligners across diverse domains. Furthermore, our analysis reveals that ensemble composition directly affects the precision–recall trade-off: heterogeneous cross-paradigm ensembles generally improve precision, whereas homogeneous LLM ensembles more often achieve higher overall F1-scores. These findings demonstrate that systematic ensemble learning offers a robust and reproducible strategy for OA while providing practical guidance for selecting ensemble compositions under different alignment scenarios.
\end{abstract}
%%
%% Keywords. The author(s) should pick words that accurately describe
%% the work being presented. Separate the keywords with commas.
  
%%
%% This command processes the author and affiliation and title
%% information and builds the first part of the formatted document.

\maketitle

\begin{itemize}[]
\item \textbf{URL}: \url{https://ontoaligner.readthedocs.io/}
\item \textbf{GitHub}: \url{https://github.com/sciknoworg/OntoAligner}
\item \textbf{PyPi}:  \url{https://pypi.org/project/OntoAligner/}
\item \textbf{License}: Apache License 2.0
\item \textbf{Experimental Resources}: \url{https://doi.org/10.5281/zenodo.21736780}
\end{itemize}

\section{Introduction}
Ontology alignment (OA) is the process of finding and defining semantic correspondences between elements of different ontologies so that data and knowledge from heterogeneous sources can interoperate accurately. The origins of OA trace back to the 1980s and 1990s, with foundations in knowledge representation, artificial intelligence (AI), and database schema matching~\cite{rahm2001survey,milo1998using}. With the emergence of the Semantic Web in the late 1990s and early 2000s, OA gained significant attention as widely adopted languages such as RDF and OWL revealed that independently developed ontologies often described similar concepts using different structures and terminologies~\cite{melnik2002similarity,PROMPT2000,noy2001anchor,doan2002learning,kalfoglou2003ontology}. Research accelerated rapidly following the inaugural International Semantic Web Conference (ISWC) in 2002, leading to standardized benchmarks via the Ontology Alignment Evaluation Initiative (OAEI) launched in 2004\footnote{\url{https://oaei.ontologymatching.org/}} and dedicated venues such as the International Workshop on Ontology Matching (OM) established in 2006\footnote{\url{http://om2006.ontologymatching.org/}}~\cite{euzenat2013ontology}. These community efforts drove advances in lexical~\cite{faria2013agreementmakerlight}, structural~\cite{jimenez2011logmap}, semantic~\cite{li2008rimom}, and instance-based~\cite{isaac2007instance} techniques. Between 2016 and 2021, the field integrated machine learning and deep learning models, particularly knowledge graph embeddings (KGEs), to capture complex semantic relationships beyond surface similarities~\cite{kolyvakis2018deepalignment,chen2021owl2vec,iyer2020veealign,he2022bertmap}. Most recently, large language models (LLMs) have opened new frontiers in handling complex and scalable alignment processes~\cite{hertling2023olala,qiang2023agent,babaei2024llms4om,babaei2025ontoaligner}.

Although OA has historically evolved through a succession of distinct methodological paradigms often treated as competing alternatives, modern frameworks such as OntoAligner~\cite{babaei_giglou_2026_21206300} aim to unify these paradigms within a modular ecosystem~\cite{giglou2025ontoaligner}. Yet, making heterogeneous aligners available through a common framework does not by itself determine how their predictions should be combined into a single, reliable alignment. OntoAligner enables lexical, structural, embedding-based, and generative aligners to operate through a shared pipeline, but a general and configurable mechanism for reconciling their complementary and sometimes conflicting outputs has remained missing. This exposes a central methodological question: \textit{how can heterogeneous alignment techniques be integrated so that their strengths reinforce one another while their individual weaknesses are mitigated?} Addressing this question is particularly important because no single paradigm consistently dominates across ontology domains. Lexical matchers can achieve high recall when ontologies are terminologically similar but may degrade under semantic drift; KGE-based aligners can effectively capture relational structures but remain sensitive to graph sparsity; and LLM-driven aligners can generalize well in open-vocabulary contexts, although their performance varies considerably across models, providers, and domains.

%Although OA has historically evolved through a succession of distinct methodological paradigms often treated as competing alternatives, modern frameworks such as OntoAligner aim to unify these paradigms into a modular ecosystem~\cite{giglou2025ontoaligner}. However, the primary bottleneck in OA is no longer the availability or maturity of individual matching paradigms. Lexical, structural, embedding-based, and generative aligners are now highly developed and readily composable; yet, the field has predominantly deferred the critical question of \textit{how to optimally integrate these diverse techniques?} Because no single paradigm achieves universal dominance across all ontology domains, their complementarity is highly significant. Lexical matchers excel in high-recall scenarios with terminologically proximal ontologies but degrade under semantic drift; KGE-based aligners effectively capture relational structures but remain sensitive to graph sparsity; and LLM-driven aligners demonstrate strong generalization in open-vocabulary contexts, yet their performance varies considerably across different models and providers.

Rather than treating this variability solely as a limitation, we view it as motivation for systematic ensemble alignment. Contemporary OA toolkits increasingly make heterogeneous alignment techniques interoperable, yet providing them within a common framework does not by itself determine how their potentially complementary and conflicting predictions should be combined. Although recent work has introduced retrieval-augmented generation (RAG)~\cite{hertling2023olala,babaei2024llms4om} and in-context reasoning into OA, LLM-based aligners have largely been studied individually, while prior work within OntoAligner~\cite{giglou2025ontoaligner} has primarily considered open-weight models. Consequently, a unified and configurable mechanism for combining conventional, KGE-based, and LLM-driven aligners, including both open-weight and API-based models, remains insufficiently investigated. To address this gap, this work makes the following central contribution:
\begin{quote}
    \textit{We introduce OntoAligner-Ensemble, a unified and configurable ensemble mechanism that combines voting-based fusion with post-fusion selection to integrate predictions from heterogeneous ontology aligners, and evaluate it across multiple domains using lightweight, KGE-based, and RAG-based aligners built on both open-weight and API-based LLMs.}
\end{quote}

We evaluate this contribution through three research questions: \textbf{RQ1:} How do lightweight, KGE-based, and RAG/LLM-based aligners compare across heterogeneous OA tasks in terms of precision, recall, and F1-score? \textbf{RQ2:} To what extent can voting-based ensemble fusion improve alignment quality and balance precision and recall relative to individual aligners and established task baselines? \textbf{RQ3:} How does ensemble composition affect performance, particularly when comparing a heterogeneous ensemble spanning multiple alignment paradigms with a homogeneous ensemble of LLM-based aligners? To investigate these questions, we evaluate five aligners and two ensemble configurations across eight tasks from five OAEI tracks spanning beyond-equivalence, circular-economy, anatomy, materials-science, and biodiversity settings. The evaluation compares conventional similarity-based, KGE-based, and RAG/LLM-based approaches, including models from open-weight and API-based ecosystems, under OA scenarios that differ in domain, scale, semantic complexity, and structural characteristics.

The remainder of this paper is organized as follows: Section~\ref{sec:formalization} formalizes the OA task;
Section~\ref{sec:relatedwork} reviews related work; Section~\ref{sec:method} presents the proposed ensemble alignment  framework; Section~\ref{sec:experiments} describes the experimental setup and evaluation methodology; Section~\ref{sec:results} discusses the empirical findings; and Section~\ref{sec:conclusion} concludes the paper and outlines future research directions.

\section{Problem Formalization}
\label{sec:formalization}

Let $C_{\text{source}} \in O_{\text{source}}$ and $C_{\text{target}} \in O_{\text{target}}$ denote concepts from the source ontology $O_{\text{source}}$ and the target ontology $O_{\text{target}}$, respectively. Formally, the OA is defined as a:
\begin{equation}
    OA(O_{\text{source}}, O_{\text{target}}) := M
\end{equation}
where
\begin{equation}
    M = \left\{ \left(s, t, r, S_r\right)
    \mid s \in O_{\text{source}},\;
    t \in O_{\text{target}}
    \right\}
\end{equation}
and $r \in \{\equiv,\sqsubseteq,\sqsupseteq\}$ denotes the semantic correspondence between the source concept $s$ and the target concept $t$, representing \emph{equivalence}, \emph{subsumption} (subclass), or \emph{inverse subsumption} (superclass), respectively. The confidence score $S_r \in [0,1]$ indicates the likelihood of the predicted correspondence.

\section{Related Work}
\label{sec:relatedwork}
OA has traditionally relied on combining lexical, structural, and semantic similarity measures using manually designed aggregation strategies~\cite{khan2023ontology}. While these approaches have shown competitive performance, their effectiveness often depends on the characteristics of the ontologies being aligned, motivating the adoption of machine learning techniques to learn better combinations of matching evidence automatically. One of the earliest machine learning approaches was proposed by Eckert et al.~\cite{5Eckert}, who introduced a meta-level learning framework that combines the outputs of multiple ontology aligners using a supervised classifier. Their results demonstrated that learned ensembles consistently outperform individual matchers and simple voting strategies. Later work of Nkisi-Orji et al.~\cite{6Nkuisi} has extended this idea by incorporating richer semantic representations. For example, Random Forest-based OA combines traditional similarity measures with word embedding features, enabling the model to learn effective combinations without manually tuning similarity weights.

Several studies have further explored ensemble learning for OA. The ROME framework~\cite{1Ding} applies Bagging and Boosting techniques to improve robustness and generalization across OA tasks. Similarly, Xue et al.~\cite{2Xue} introduced a collaborative OA based on dual-population genetic programming that employs active meta-learning and a Random Forest meta-classifier to construct high-quality similarity features and improve alignment accuracy. Interactive approaches such as DualLoop~\cite{7Cheng} also adopt ensemble principles by combining multiple heuristic matchers with active learning to reduce human annotation effort. 

Ensemble techniques have also been applied in domain-specific settings. For instance, biomedical OA has benefited from integrating word embedding similarities with existing ontology aligners~\cite{4dearing2017exploring}, demonstrating that learned semantic features complement traditional matching methods. Although focusing on instance matching,  \cite{gharpure2024hybrid} introduced a hybrid ensemble classifier for identifying equivalent entities across heterogeneous datasets, providing insights that are also applicable to ontology matching and semantic data integration.

Overall, these studies demonstrate that ensemble learning can enhance OA by effectively combining complementary similarity measures, matching strategies, and alignment paradigms. However, despite its potential, ensemble-based approaches remain relatively underexplored within the broader history of OA research. One possible explanation is the additional computational overhead introduced by ensemble strategies, as combining multiple aligners typically increases execution time and resource requirements. While this trade-off may limit their applicability in time-sensitive scenarios, it can be justified in alignment tasks where accuracy, robustness, and the ability to handle heterogeneous ontologies are prioritized over response efficiency. Motivated by this observation, our work builds upon existing ensemble-based OA research by investigating a unified ensemble framework that integrates multiple ontology aligners and leverages their complementary strengths to improve alignment quality across diverse OA scenarios.

\section{OntoAligner-Ensemble Framework}
\label{sec:method}
% add material in the zendo as well.

\begin{figure}[t]
    \centering
    \includegraphics[width=\linewidth]{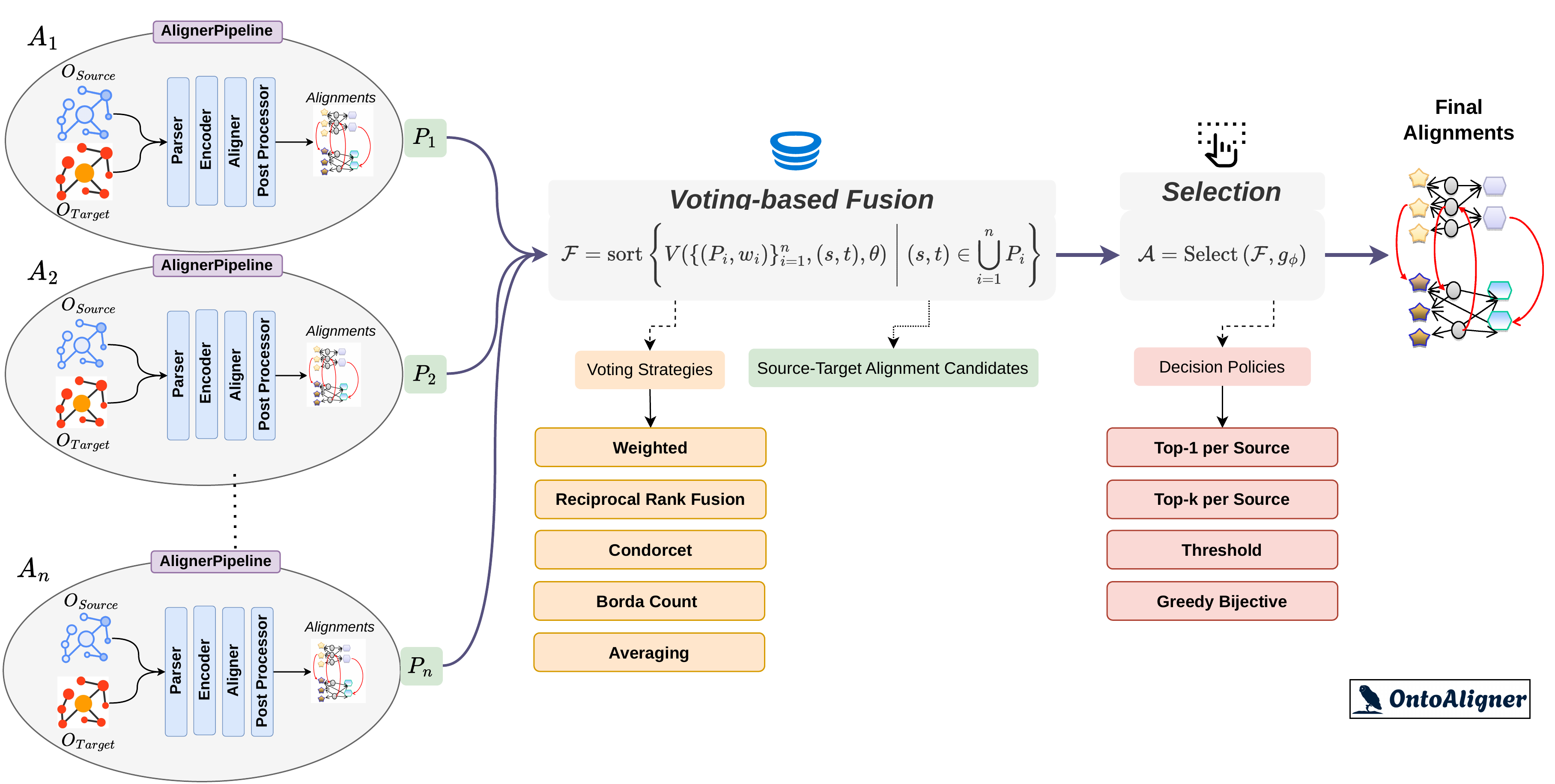}
    \caption{Overview of the OntoAligner-Ensemble architecture. Individual aligner pipelines ($A_1 \dots A_n$) process source ($O_{Source}$) and target ($O_{Target}$) ontologies through parsing, encoding, alignment, and post-processing steps. Candidate alignments ($P_i$) are combined using a configurable Voting-based Fusion layer—employing one of the strategies such as Weighted, Reciprocal Rank Fusion, Condorcet, Borda Count, or Averaging. Final alignments ($\mathcal{A}$) are generated after applying candidate selection and decision policies (e.g., Top-1, Top-$k$, Threshold, or Greedy Bijective).}
    \label{fig:framework}
\end{figure}

Due to the complexity and diversity of real-world ontological structures, individual aligner approaches might capture only specific aspects of semantic similarity, such as lexical information, structural characteristics, or contextual knowledge. To address this limitation, we propose the OntoAligner-Ensemble framework that combines multiple independent alignment pipelines through a configurable voting-based aggregation mechanism.  The proposed framework is illustrated in \autoref{fig:framework} and follows a two-stage decision process: 1) \underline{\textit{Fusion stage}}, multiple aligners generate candidate correspondences, which are aggregated using a voting strategy to obtain a unified ranking of candidate alignments. 2) \underline{\textit{Selection stage}}, a post-fusion selection strategy converts the ranked candidate space into the final alignment by applying task-specific constraints.

The framework is designed to be independent of the underlying alignment algorithms. Any ontology aligner capable of producing source--target correspondence predictions can be incorporated as a constituent aligner; all these capabilities are possible because of the high-level abstraction of \texttt{AlignerPipeline} module~\footnote{See \texttt{AlignerPipeline} usage at \url{https://ontoaligner.readthedocs.io/developerguide/pipeline.html}}that is responsible for running the aligner steps, including parsing, encoding, aligner, and post-processing (if necessary), all of which are according to the OntoAligner backbone architecture. In the following, we describe the principal components of OntoAligner-Ensemble.

\paragraph{Ensemble Representation.} Let an ensemble consist of $n$ independent aligners $A = \{A_1, A_2, ..., A_n\}$, where each aligner $A_i$ corresponds to an individual \texttt{AlignerPipeline}. Each aligner produces a set of candidate correspondences: $P_i = \{(s, t, q_i(s, t))\}$, where $s \in O_{source}$, $t \in O_{target}$ and $q_i(s,t)$ is the confidence score that $A_i$ assigns to the candidate correspondence $(s,t)$. To account for differences in reliability among aligners, each aligner is assigned a weight $w_i \in \mathbb{R}$ (the default is set to 1), where larger values indicate a greater contribution during the fusion process. Therefore, the complete ensemble input is represented as:
\begin{equation}
\mathcal{P} = \{ (P_i, w_i)\}_{i=1}^n
\end{equation}
Once the predictions have been generated by the $n$ aligners, the framework converts them into a unified representation because different aligners may use different output formats. For example, the output of a retrieval-based aligner is converted from $s \mapsto \{(t_1,q_1),(t_2,q_2),\ldots\}$ to individual triples $(s,t_j,q_j)$. This allows different aligners to participate in the same ensemble process. Additionally, duplicate $(s, t)$ are removed by retaining the highest-ranked occurrence according to the confidence score: $(s, t, q) = argmax q(s,t)$. This guarantees a unique candidate representation before voting.

\paragraph{Voting-Based Fusion.}
The fusion component aggregates predictions from multiple constituent aligners. Let
\begin{equation}
    \mathcal{C}
    =
    \bigcup_{i=1}^{n}
    \left\{
        (s,t)
        \;\middle|\;
        (s,t,q_i(s,t)) \in P_i
    \right\}
\end{equation}
denote the set of candidate source--target pairs proposed by at least one aligner. For each candidate $(s,t) \in \mathcal{C}$, the fused score is defined as
\begin{equation}
    S(s,t)
    =
    \mathcal{V}
    \left(
        \{(P_i,w_i)\}_{i=1}^{n},
        (s,t),
        \theta
    \right)
\end{equation}
where $\mathcal{V}$ denotes the selected voting strategy and $\theta$ contains its strategy-specific parameters. The ranked fused prediction set is then defined as
\begin{equation}
    \mathcal{F}
    =
    \operatorname{sort}_{\downarrow S}
    \left(
        \left\{
            (s,t,S(s,t))
            \;\middle|\;
            (s,t) \in \mathcal{C}
        \right\}
    \right)
\end{equation}
where candidates are ordered by decreasing fused score. OntoAligner-Ensemble supports weighted voting, Reciprocal Rank Fusion (RRF), Condorcet voting, Borda count, and score averaging.

%\paragraph{Voting-Based Fusion.} The fusion component aggregates predictions from multiple aligners. Instead of assuming that one aligner is always optimal, the ensemble estimates the collective support for each candidate correspondence. Let $(s, t) \in \cup_{i=1}^A P_i$ be the candidate alignments appearing in at least one aligner.  The fused score is defined as $S(s,t) = \mathbf{V}(\{ (P_i, w_i)\}_{i=1}^A, \theta)$, where $\mathbf{V}$ represents the voting strategy and $\theta$ denotes strategy-specific parameters. Therefore, the fused prediction set is obtained as $\mathcal{F} = {(s, t, S(s,t)}$.  Different voting strategies define different implementations of $\mathbf{V}$. OntoAligner supports the weighted, Reciprocal Rank Fusion (RRF), Condorcet, Borda count, and averaging-based voting strategies.

\paragraph{Post-Fusion Selection.} After fusion, the framework contains a ranked candidate pool $\mathcal{F}$. A selection operator converts this candidate pool into the final alignment of $\mathcal{A}= Select(\mathcal{F}, g_\phi)$, where $\phi$ represents the selection strategy parameters. Moreover, $g_\phi$ is defined as a decision policy that is responsible for selecting appropriate alignments based on its own selection policy; the default for $g_\phi$ is none. However, there are four different decision policies which are defined as follows:
\begin{itemize}
    \item \textbf{Top-1 per Source Selection}. This decision policy aims to select the highest scoring target for each source: $t^* = argmax S(s, t)$.

    \item \textbf{Top-$k$ per Source Selection}. For each source entity, the framework retains the top-$k$ highest-scoring target candidates. Using this policy, optionally, candidates are first filtered using a relative score margin $S(s,t) \geq m \cdot \max_{t^\prime} S(s,t^\prime)$, where $m \in [0,1]$ is the margin parameter. When both constraints are specified, only the top-$k$ candidates satisfying the margin criterion are retained.

    \item \textbf{Threshold-based Selection}. Only correspondences exceeding a confidence threshold are retained: $A = \{ (s, t) \in \mathcal{F} | S(s,t) \geq \gamma \}$, where $\gamma$ is the confidence threshold.
    
    \item \textbf{Greedy Bijective Selection}. For applications requiring one-to-one alignment, this decision policy applies a greedy bijective selection strategy as follows: $\forall (s_i, t_i), (s_j, t_j) \in \mathcal{A} $, where $s_i \neq s_j \wedge t_i \neq t_j$, meaning that each source and target entity can appear at most once.
\end{itemize}

This modular formulation enables OntoAligner-Ensemble to integrate heterogeneous OA systems within a unified framework while supporting configurable voting strategies and post-fusion selection policies that can be adapted to different datasets and application requirements~\footnote{See \texttt{EnsembleLearningAligner} usage at \url{https://ontoaligner.readthedocs.io/aligner/ensemble_learning.html}}.

\section{Experimental Setup}
\label{sec:experiments}

\begin{table}
    \centering
    \caption{OAEI tracks and tasks statistics across source, target, and alignments.}
    \label{exp-datasets}
    \resizebox{\textwidth}{!}{ 
    \begin{tabular}{llccc}
        \hline
         \textbf{Track} & \textbf{Task} & \textbf{$O_{source}$ Classes}& \textbf{$O_{target}$ Classes} & \textbf{References} \\
        \hline
        \multirow{3}{*}{\textit{Beyond Equivalence}} & G1 - Web & 727  &  1,132 & 339  \\
                                                       & G2 - Diseases &  1,108 & 5,145  & 355  \\
                                                       & G3 - Text & 334  &  259 &  762 \\
        \hline
        \multirow{2}{*}{\textit{Circular Economy}} & CEON - BiOnto & 228 & 779 & 29 \\
                                                   & CEON - MatOnto & 228  &  846 & 16  \\
        \hline
         \textit{Anatomy} & Mouse-Human & 2,743 & 3,304  & 1,516 \\
         \hline
         \textit{Material Science and Engineering} & MI--MatOnto & 545 & 847 & 302 \\
         \hline
         \textit{Biodiversity and Ecology} &Fish--Zooplankton &  145  & 56  & 15 \\
         \hline
    \end{tabular}
    }
\end{table}
\paragraph{Evaluation Datasets: OAEI Tracks and Tasks.} We selected eight OA tasks from five tracks of the OAEI campaign to evaluate OntoAligner-Ensemble across diverse domains. The statistics of the selected datasets, including the number of source ontology classes, target ontology classes, and reference correspondences, are summarized in \autoref{exp-datasets}. The selected tracks include: \textit{Beyond Equivalence}~\cite{hertling2023transformer,arnold2014enriching} (G1--Web, G2--Diseases, and G3--Text), \textit{Circular Economy}~\cite{blomqvist2023cross} (CEON--BiOnto and CEON--MatOnto), \textit{Anatomy}~\cite{anatomy} (Mouse--Human), \textit{Material Science and Engineering}~\cite{mse} (MI--MatOnto), and \textit{Biodiversity and Ecology}~\cite{biodiversity} (Fish--Zooplankton). These datasets were selected to provide a comprehensive evaluation across heterogeneous domains, ontology sizes, and matching difficulties. 

The \textit{Beyond Equivalence} track evaluates OA systems beyond simple equivalence detection by considering diverse semantic relations, including equivalence, subsumption, overlap, and disjointness. It contains heterogeneous benchmarks from industrial classification schemes and the STROMA/TaSeR repository. In this work, we use the STROMA/TaSeR-based tasks (G1--Web, G2--Diseases, and G3--Text), which provide diverse matching scenarios with varying ontology sizes and semantic complexity. The \textit{Circular Economy} and \textit{Material Science and Engineering} tracks evaluate domain-specific ontologies with relatively smaller alignment spaces, while the \textit{Anatomy} track represents a larger matching task with thousands of ontology classes and reference correspondences. Finally, the \textit{Biodiversity and Ecology} task provides a small-scale evaluation scenario with limited reference alignments. Overall, the selected benchmarks cover a broad range of task characteristics, including small-scale tasks (Fish--Zooplankton, CEON--BiOnto, and CEON--MatOnto), medium-scale tasks (G1--Web, G3--Text, and MI--MatOnto), and large-scale tasks (G2--Diseases and Mouse--Human). This diversity enables a comprehensive assessment of the robustness and generalization capability of both individual OA systems and the proposed ensemble framework.

\paragraph{Individual Aligners.}To evaluate both individual OA systems and their ensemble combinations, we selected five aligners representing three complementary categories: \emph{Lightweight}, \emph{KGE}~\cite{giglou2025ontoaligner}, and \emph{RAG}~\cite{babaei2024llms4om}. All aligners were implemented using the \texttt{AlignerPipeline} interface provided by OntoAligner, enabling a unified execution and prediction format across heterogeneous alignment approaches.
\begin{itemize}
    \item \textit{Lightweight.} We employed the fuzzy string matching aligner available in OntoAligner\footnote{\url{https://ontoaligner.readthedocs.io/aligner/lightweight.html}}, using a similarity threshold of \textit{0.7} for all experiments.

    \item \textit{KGE.} We selected the ConvE-based aligner with the same hyperparameter configuration reported in our previous work~\cite{giglou2025ontoaligner}.

    \item \textit{RAG.} To evaluate the robustness of the proposed ensemble across different LLM ecosystems, we considered three RAG configurations:
    \begin{itemize}
        \item \textit{Qwen.} Qwen3.5-9B~\cite{qwen3.5} as the generator and Qwen3-Embedding-4B~\cite{qwen3embedding} as the retriever, both from the Qwen model family.
        \item \textit{GPT.} GPT-5.4-Nano~\footnote{\url{https://developers.openai.com/api/docs/models/gpt-5.4-nano}} as the generator and \texttt{text-embedding-3-small}~\footnote{\url{https://developers.openai.com/api/docs/models/text-embedding-3-small}} as the retriever, representing the OpenAI model family.
        \item \textit{Gemini.} The Gemini 2.5 Flash-Lite~\footnote{\url{https://ai.google.dev/gemini-api/docs/models/gemini-2.5-flash-lite}} as the generator and  EmbeddingGemma-300m~\cite{embedding_gemma_2025} as the retriever, representing the Google model family.
    \end{itemize}
\end{itemize}
The hyperparameter values used throughout the experiments were selected manually based on preliminary empirical observations and prior experience with the underlying alignment methods. No systematic hyperparameter optimization was performed, as the large configuration space arising from multiple aligners, voting strategies, and selection policies would make exhaustive or automated tuning computationally prohibitive. 

\paragraph{Ensemble Configurations.} To evaluate the effectiveness of ensemble-based aligners, we constructed two configurations of OntoAligner-Ensemble. Each ensemble treats individual aligners as independent techniques and combines their predictions through a voting-based fusion strategy followed by a decision policy.
\begin{itemize}
    \item \textit{Ens. (All)}. This configuration combines all available aligners across different categories, including the lightweight string-based aligner, the KGE-based aligner, and the three RAG-based aligners. This configuration evaluates whether heterogeneous alignment systems with complementary characteristics can improve overall alignment performance through collective decision-making.
    \item \textit{Ens. (LLMs)}. This configuration focuses exclusively on LLM-based aligners and combines the three RAG aligners, namely Qwen, GPT, and Gemini. This setting investigates whether an ensemble of different LLM-based aligners can provide more robust predictions compared with individual LLM aligners.
\end{itemize}
For all ensemble experiments, each aligner was assigned an equal weight ($w_i=1.0$). The majority voting strategy (supported by weighted voting) was employed with a minimum valid vote of 3 for \textit{Ens. (All)} and 2 for  \textit{Ens. (LLMs)}. After fusion, the resulting ranked correspondence candidates were processed using the \textit{top-1-source} decision policy to obtain the final alignment set. This configuration was selected to provide a consistent comparison across datasets while avoiding additional tuning of ensemble-specific parameters. Nevertheless, the proposed framework is fully configurable, allowing practitioners to adjust both model-specific and ensemble-level hyperparameters to suit different datasets, computational budgets, and application requirements.

\section{Results and Discussion}
\label{sec:results}

We evaluate the five individual aligners and the two proposed ensemble configurations, \textit{Ens. (All)} and \textit{Ens. (LLMs)}, across eight benchmark tasks from five OAEI tracks. The experimental results, measured in terms of precision, recall, and F1-score, are summarized in \autoref{tab:results}. The remainder of this section is organized around the three research questions: RQ1 examines the performance profiles of the individual aligners, RQ2 evaluates the effectiveness of the proposed voting-based ensemble, and RQ3 investigates how ensemble composition affects performance across domains and task types.

\begin{table}[t]
\centering
\caption{Ontology matching results across all evaluation datasets. ``\textit{Base.}'' denotes the baseline aligner for each dataset. Best precision, recall, and F1-score for each task are shown in bold.}
\label{tab:results}
% \scriptsize
% \setlength{\tabcolsep}{3pt}
% \renewcommand{\arraystretch}{1.05}
\resizebox{\textwidth}{!}{ 
\begin{tabular}{lll cccccccc}
\toprule
\multirow{2}{*}{\textbf{Task}} &
\multirow{2}{*}{\textbf{Base. Aligner}} &
\multirow{2}{*}{\textbf{Metric}} &
\multicolumn{8}{c}{\textbf{Aligners}} \\
\cline{4-11}
& & &
\textbf{Base.} &
\textbf{Fuzzy} &
\textbf{KGE} &
\textbf{Qwen} &
\textbf{GPT-5.4} &
\textbf{Gemini} &
\textbf{Ens. (All)} &
\textbf{Ens. (LLMs)} \\
\midrule

\multirow{3}{*}{G1-Web} & \multirow{3}{*}{MDMapper~\cite{liu2024mdmapper}}
& Precision & \textbf{88.2} & 39.8 & 68.9 & 65.5 & 56.3 & 61.7 & 74.2 & 68.7 \\
& & Recall    & 36.3 & 53.0 & 41.2 & 51.0 & \textbf{54.8} & 51.9 & 48.3 & 51.3 \\
& & F1         & 51.5 & 45.5 & 51.6 & 57.3 & 55.6 & 56.4 & 58.5 & \textbf{58.7} \\
\midrule

\multirow{3}{*}{G2-Diseases}  & \multirow{3}{*}{LogMapBio~\cite{jimenez2011logmap}}
& Precision & \textbf{60.1} & 32.5 & \textbf{60.1} & 50.1 & 48.6 & 51.2 & 55.2 & 54.6 \\
& & Recall    & 61.0 & 69.2 & 54.3 & 72.9 & \textbf{75.7} & 71.5 & 69.4 & 74.9 \\
& & F1         & 60.5 & 44.3 & 57.1 & 59.4 & 59.2 & 59.6 & 61.4 & \textbf{63.1} \\
\midrule

\multirow{3}{*}{G3-Text}  & \multirow{3}{*}{LogMap~\cite{jimenez2011logmap}}
& Precision & 43.7 & 40.0 & 36.7 & \textbf{48.3} & 41.7 & 45.2 & 45.9 & 46.9 \\
& & Recall    & 7.3 & \textbf{9.4} & 5.1 & 9.0 & 8.9 & 8.7 & 8.13 & 9.0  \\
& & F1         & 12.5  & \textbf{15.2} & 8.9  & 15.0  & 14.7  & 14.7  & 13.8 & \textbf{15.2} \\
\midrule

\multirow{3}{*}{CEON--BiOnto}  & \multirow{3}{*}{LogMap~\cite{jimenez2011logmap}}
& Precision & 67.6 & 32.9 & \textbf{78.2} & 47.4 & 40.2 & 51.7 & 63.6 & 55.7 \\
& & Recall    & 86.2 & 93.1 & 62.0 & 96.5 & \textbf{100} & \textbf{100} & 96.5 & \textbf{100} \\
& & F1         & 75.8 & 48.6 & 69.2 & 63.6 & 57.4 & 68.2 & \textbf{76.7} & 71.6 \\
\midrule

\multirow{3}{*}{CEON--MatOnto}  & \multirow{3}{*}{LogMap~\cite{jimenez2011logmap}}
& Precision & \textbf{35.6} & 22.8 & 33.3 & 27.7 & 27.7 & 28.0 & 33.3 & 29.4 \\
&& Recall    & \textbf{100} & \textbf{100} & 81.2 & 93.7 & 93.7 & \textbf{100} & 93.7 & 93.7 \\
& & F1         & \textbf{52.5} & 37.2 & 47.2 & 42.8 & 42.8 & 43.8 & 49.1 & 44.7 \\
\midrule

\multirow{3}{*}{Fish--Zooplankton}  & \multirow{3}{*}{LogMapLt~\cite{jimenez2011logmap}}
 & Precision & 80.0 & 64.2 & \textbf{100} & 92.8 & 93.3 & 100 & 100 & \textbf{100} \\
& & Recall    & 53.3 & 60.0 & 53.3 & 86.6 & \textbf{93.3} & 86.6 & 73.3 & \textbf{93.3} \\
& & F1         & 64.0 & 62.0 & 69.5 & 89.6 & \textbf{93.3} & 92.8 & 84.6 & \textbf{96.5} \\
\midrule

\multirow{3}{*}{MI--MatOnto}  & \multirow{3}{*}{Matcha~\cite{cotovio2024matcha}}
& Precision & 75.6 & 39.3 & \textbf{93.3} & 61.7 & 55.0 & 63.5 & \textbf{84.3} & 72.7 \\
& & Recall    & 21.9 & 21.5 & 9.2 & \textbf{32.1} & 27.1 & 29.4 & 23.1 & 29.1 \\
& & F1         & 33.9 & 27.8 & 16.8 & \textbf{42.2} & 36.3 & 40.2 & 36.3 & 41.6 \\
\midrule

\multirow{3}{*}{Mouse--Human}  & \multirow{3}{*}{Matcha~\cite{cotovio2024matcha}}
& Precision & 95.1 & 53.8 & 99.0 & 86.9 & 87.0 & 89.4 & \textbf{96.5} & 93.0 \\
& & Recall    & \textbf{93.1} & 76.7 & 61.3 & 90.6 & 89.3 & 88.3 & 85.5 & 90.3 \\
& & F1         & \textbf{94.1} & 63.2 & 75.6 & 88.7 & 88.1 & 88.8 & 90.6 & 91.6 \\
\bottomrule
\end{tabular}
}
\end{table}

\subsection{RQ1: Performance of Individual Aligners}
\label{sec:results-rq1}

No individual aligner consistently achieves the best performance across all evaluation tasks. Instead, the strongest approach varies with the benchmark, indicating that OA remains task-dependent. The three evaluated aligner categories exhibit distinct precision--recall profiles.

\begin{itemize}
    \item \textbf{Lightweight.} The fuzzy string aligner achieves high recall on tasks with substantial lexical overlap, including \textit{CEON--BiOnto} (93.1\%) and \textit{CEON--MatOnto} (100.0\%). However, this behavior is accompanied by comparatively low precision, such as 32.5\% on \textit{G2--Diseases} and 22.8\% on \textit{CEON--MatOnto}. These results indicate that reliance on surface-form similarity can retrieve many candidate correspondences while also introducing false positives.

    \item \textbf{KGE-based.} The ConvE-based KGE aligner achieves high precision on several tasks, consistent with observations in prior work~\cite{giglou2025ontoaligner}. It reaches 100.0\% precision on \textit{Fish--Zooplankton} and 93.3\% on \textit{MI--MatOnto}. This precision-oriented behavior is nevertheless accompanied by low recall on some tasks, including 9.2\% on \textit{MI--MatOnto} and 5.1\% on \textit{G3--Text}.

    \item \textbf{RAG/LLM-based.} The RAG-based aligners generally provide broader correspondence coverage, although their performance varies across models and tasks. On \textit{MI--MatOnto}, Qwen achieves an F1-score of 42.2\%, outperforming the established baseline and all other individual aligners. On \textit{Fish--Zooplankton}, GPT-5.4-Nano achieves a balanced precision and recall of 93.3\%. The three LLM-based aligners nevertheless exhibit different performance profiles; for example, the precision of GPT-5.4-Nano ranges from 40.2\% on \textit{CEON--BiOnto} to 93.3\% on \textit{Fish--Zooplankton}. Thus, no individual LLM is consistently superior across all tasks.
\end{itemize}

For six of the eight tasks, all three RAG/LLM-based aligners achieve higher recall than the corresponding established baseline. The exceptions are \textit{CEON--MatOnto}, where the baseline already achieves 100.0\% recall, and \textit{Mouse--Human}, where the baseline recall of 93.1\% exceeds that of the three RAG aligners. The recall improvement is particularly pronounced on \textit{Fish--Zooplankton}, where GPT-5.4-Nano reaches 93.3\% recall compared with 53.3\% for LogMapLt.

The broader coverage of the RAG/LLM-based aligners is often accompanied by lower precision than that of the established baselines. Conversely, the KGE aligner frequently achieves high precision but lower recall. The lightweight aligner can also obtain high recall in lexically similar tasks, although often with a substantial precision reduction. These complementary performance profiles provide the motivation for the ensemble analysis in RQ2.

Established task baselines nevertheless remain competitive. They achieve the highest F1-score on \textit{CEON--MatOnto} and \textit{Mouse--Human} and remain close to the best-performing approach on tasks such as \textit{G2--Diseases} and \textit{CEON--BiOnto}. Overall, RQ1 shows that the relative effectiveness of lightweight, KGE-based, and RAG/LLM-based aligners depends on the task: lightweight and RAG-based approaches generally favor recall, whereas the evaluated KGE aligner generally favors precision, and no individual alignment paradigm dominates across all benchmarks.

\subsection{RQ2: Effectiveness of the Proposed Voting-Based Ensemble}
\label{sec:results-rq2}

The results indicate that voting-based fusion can improve the balance between precision and recall, although its effectiveness remains task-dependent. At least one ensemble configuration achieves or ties for the highest F1-score on five of the eight tasks: \textit{G1--Web}, \textit{G2--Diseases}, \textit{G3--Text}, \textit{CEON--BiOnto}, and \textit{Fish--Zooplankton}. These results show that combining predictions from multiple aligners can improve upon the constituent systems when their predictions provide complementary alignment evidence.

The clearest improvements are observed on \textit{G1--Web}, \textit{G2--Diseases}, and \textit{Fish--Zooplankton}. On \textit{G1--Web}, \textit{Ens. (LLMs)} achieves the highest F1-score of 58.78\%, exceeding the best individual aligner, Qwen, at 57.37\%, and the MDMapper baseline at 51.50\%. On \textit{G2--Diseases}, \textit{Ens. (LLMs)} reaches 63.1\% F1, compared with 59.6\% for the strongest individual aligner, Gemini, and 60.5\% for the LogMapBio baseline. The largest ensemble gain occurs on \textit{Fish--Zooplankton}, where \textit{Ens. (LLMs)} obtains 96.5\% F1, improving upon the strongest individual aligner, GPT-5.4-Nano, at 93.3\%, and the LogMapLt baseline at 64.0\%.

The ensemble configurations also provide competitive results on \textit{G3--Text} and \textit{CEON--BiOnto}. On \textit{G3--Text}, \textit{Ens. (LLMs)} reaches 15.2\% F1, tying the fuzzy string aligner for the highest result and exceeding the LogMap baseline at 12.5\%. On \textit{CEON--BiOnto}, \textit{Ens. (All)} achieves the highest overall F1-score of 76.7\%, slightly surpassing the LogMap baseline at 75.8\% and outperforming all individual aligners.

However, voting-based fusion does not uniformly outperform the strongest individual method or established baseline. On \textit{MI--MatOnto}, Qwen achieves the highest F1-score of 42.2\%, compared with 41.6\% for \textit{Ens. (LLMs)} and 36.3\% for \textit{Ens. (All)}. On \textit{CEON--MatOnto}, the LogMap baseline remains strongest at 52.5\% F1, whereas \textit{Ens. (All)} and \textit{Ens. (LLMs)} achieve 49.1\% and 44.7\%, respectively. Similarly, on \textit{Mouse--Human}, the Matcha baseline retains the highest F1-score of 94.1\%, followed by \textit{Ens. (LLMs)} at 91.6\% and \textit{Ens. (All)} at 90.6\%.

The results therefore show that voting-based fusion is most effective when agreement among the constituent aligners suppresses individual errors without removing too many valid correspondences. When one constituent aligner is already substantially stronger than the others, or when voting reduces recall too aggressively, the ensemble may remain below the strongest individual method or specialized task baseline.

Ensembling also introduces an expected computational trade-off because the predictions of multiple constituent aligners must be generated and combined. Runtime was not evaluated as an experimental metric in this study, so no quantitative efficiency comparison can be made. Nevertheless, the modular implementation within OntoAligner allows the number and types of constituent aligners to be adjusted according to task requirements and the available computational budget. Overall, RQ2 shows that the proposed voting-based ensemble can improve alignment quality and the precision--recall balance, but its benefits depend on the benchmark and the relative strengths of its constituent aligners.

\subsection{RQ3: Effect of Ensemble Composition across Domains}
\label{sec:results-rq3}

The two ensemble configurations exhibit distinct precision--recall profiles. The heterogeneous configuration, \textit{Ens. (All)}, combines the lightweight, KGE-based, and three RAG/LLM-based aligners, whereas the homogeneous configuration, \textit{Ens. (LLMs)}, combines only the three RAG/LLM-based aligners. Across the eight tasks, \textit{Ens. (All)} generally favors precision, while \textit{Ens. (LLMs)} more often preserves recall. Consequently, \textit{Ens. (LLMs)} achieves a higher F1-score than \textit{Ens. (All)} on six tasks, whereas \textit{Ens. (All)} performs better on the two Circular Economy tasks.

\begin{itemize}
    \item \textbf{Homogeneous LLM ensembling (\textit{Ens. (LLMs)}).} Combining Qwen, GPT-5.4-Nano, and Gemini through majority voting achieves or ties for the highest F1-score on four tasks: \textit{G1--Web} (58.78\%), \textit{G2--Diseases} (63.1\%), \textit{G3--Text} (15.2\%), and \textit{Fish--Zooplankton} (96.5\%). The configuration outperforms every standalone LLM on \textit{G1--Web}, \textit{G2--Diseases}, and \textit{Fish--Zooplankton}. Its comparatively strong recall suggests that agreement among the LLM-based aligners retains broader semantic coverage than the more heterogeneous ensemble.

    \item \textbf{Heterogeneous multi-paradigm ensembling (\textit{Ens. (All)}).} Combining lightweight, KGE-based, and RAG/LLM-based aligners produces a more precision-oriented configuration. On \textit{Mouse--Human}, \textit{Ens. (All)} achieves 96.5\% precision, exceeding all three individual LLM aligners, although the KGE aligner remains higher at 99.0\%. On \textit{MI--MatOnto}, \textit{Ens. (All)} reaches 84.3\% precision, which is 22.6 percentage points higher than the best individual LLM precision of 61.7\% obtained by Qwen. Its recall on this task is only 23.1\%, however, resulting in an F1-score of 36.3\%. The heterogeneous ensemble therefore acts as a comparatively conservative filter: agreement across different alignment paradigms can increase precision, but may also remove valid correspondences that are supported by only a subset of the constituent aligners.
\end{itemize}

The effect of ensemble composition also differs across domains and task types:
\begin{itemize}
    \item \textbf{Biomedical and standardized domains (\textit{Mouse--Human}, \textit{G2--Diseases}).} On \textit{Mouse--Human}, the specialized Matcha baseline retains the highest F1-score of 94.1\%. Nevertheless, \textit{Ens. (LLMs)} achieves a competitive 91.6\% F1 in the reported setup without task-specific fine-tuning, while \textit{Ens. (All)} reaches 90.6\%. The heterogeneous ensemble obtains higher precision than the LLM ensemble (96.5\% versus 93.0\%), whereas the LLM ensemble preserves higher recall (90.3\% versus 85.5\%). On \textit{G2--Diseases}, \textit{Ens. (LLMs)} achieves 63.1\% F1 and outperforms both \textit{Ens. (All)} at 61.4\% and the LogMapBio baseline at 60.5\%. These results indicate that the LLM-only composition can provide competitive coverage of biomedical terminology, although the advantage is not uniform across biomedical tasks.

    \item \textbf{Materials science and Circular Economy (\textit{MI--MatOnto}, \textit{CEON--BiOnto}, \textit{CEON--MatOnto}).} On \textit{MI--MatOnto}, Qwen achieves the highest F1-score of 42.2\%, followed closely by \textit{Ens. (LLMs)} at 41.6\%, both exceeding the Matcha baseline at 33.9\% and the KGE aligner at 16.8\%. In contrast, \textit{Ens. (All)} achieves substantially higher precision (84.3\%) but lower recall (23.1\%), resulting in 36.3\% F1. This illustrates how the heterogeneous composition can favor conservative correspondences at the expense of broader coverage.

    On \textit{CEON--BiOnto}, \textit{Ens. (All)} achieves the highest F1-score of 76.7\%, exceeding the LogMap baseline at 75.8\% and \textit{Ens. (LLMs)} at 71.6\%. The LLM ensemble reaches 100.0\% recall, but its lower precision of 55.7\% reduces its overall F1-score. On \textit{CEON--MatOnto}, neither ensemble exceeds the LogMap baseline at 52.5\% F1. Nevertheless, \textit{Ens. (All)} again outperforms \textit{Ens. (LLMs)} in F1-score (49.1\% versus 44.7\%) because of its higher precision (33.3\% versus 29.4\%), while both configurations obtain the same recall of 93.7\%. The two CEON tasks therefore provide the clearest cases in which cross-paradigm agreement improves the precision--recall balance relative to the LLM-only composition.

    \item \textbf{Non-equivalence and complex mapping (\textit{G3--Text}).} Recall remains below 10\% for every evaluated method on \textit{G3--Text}, indicating that this task remains challenging for both individual and ensemble approaches. In this setting, \textit{Ens. (LLMs)} achieves 15.2\% F1, tying the fuzzy string aligner for the highest result and exceeding both \textit{Ens. (All)} at 13.8\% and the LogMap baseline at 12.5\%. The LLM-only composition therefore provides a modest advantage over the heterogeneous ensemble under this difficult mapping setting, although the low absolute recall indicates substantial remaining room for improvement.

    \item \textbf{Biodiversity and ecology (\textit{Fish--Zooplankton}).} Both ensemble configurations achieve 100.0\% precision, but their recall differs substantially. \textit{Ens. (LLMs)} reaches 93.3\% recall and 96.5\% F1, whereas \textit{Ens. (All)} obtains 73.3\% recall and 84.6\% F1. The lower recall of the heterogeneous ensemble suggests that requiring agreement across substantially different alignment paradigms can be overly restrictive for this task. Here, the homogeneous LLM composition retains more valid correspondences while preserving perfect precision.
\end{itemize}
Overall, RQ3 shows that ensemble composition directly affects the precision--recall trade-off. The heterogeneous multi-paradigm ensemble generally provides stronger precision and performs best on the two Circular Economy tasks, whereas the homogeneous LLM ensemble generally retains more recall and achieves the higher F1-score on six of the eight tasks. Neither composition is uniformly preferable; their relative effectiveness depends on the domain, the task characteristics, and the degree of agreement among the constituent aligners.

\section{Discussion and Conclusion}
% \section{Conclusion and Future Directions}
\label{sec:conclusion}

% \paragraph{Industry Perspective.} Better data products for people and AI agents. The ability to align and integrate proprietary and public ontologies describing data products with quality and on scale substantially increases the  discoverability, interpretability, trust and use of the underlying data.  Semantic interoperability is a key requirement in regulatory reporting use cases, publishing of reference data products, managerial and operational self-service analytics. Semantically enriched metadata reduces the friction associated with using data products for both humans and machines. The aligned ontological understanding enables customers to discover, understand, trust and use the published data within their own systems - simplifying the consumption, distribution, and AI readiness of the data products. Data assets which are created for one business purpose can be reused and integrated with others, on the basis of their model, by the creation of a semantic layer.
\paragraph{Industry Perspective.}  From an industrial application perspective, scaling the alignment and integration of proprietary and domain-standard ontologies directly supports the deployment of reliable data products for both human analysts and automated agents. Achieving robust semantic interoperability is crucial in regulatory reporting, reference data integration, and enterprise analytics, where data assets created for distinct business purposes must be harmonized into unified semantic layers. In these settings, raw individual aligners often face operational trade-offs: lexical aligners can fail under proprietary terminology changes, while LLM-driven aligners introduce variance across model providers. The OntoAligner-Ensemble mechanisms developed in this work help mitigate these single-model risks by providing predictable, high-precision candidate alignments. Improving the precision--recall balance through systematic fusion lowers the manual verification cost needed to build trusted data products, ultimately facilitating smoother data consumption, discovery, and inter-organizational system integration.

\paragraph{Limitations and Future Directions.} While our OA definition considers four matching relations (\textit{equivalence}, \textit{subclass}, \textit{superclass}, and \textit{disjointness}), the current framework reduces output decisions to binary classification (\textit{match} vs. \textit{no-match}). Consequently, while a positive \textit{match} implicitly can encompass equivalence, subclass, and superclass associations, the framework does not differentiate between these distinct relation types. This limitation stems from the current fusion functions, which prioritize consensus over fine-grained semantic categorization. Extending the decision layer to distinguish specific semantic relations represents a key direction for future developments of OntoAligner.

\paragraph{Conclusion.} In this work, we presented OntoAligner-Ensemble, a modular, aligner-agnostic framework designed to systematically integrate complementary candidate predictions across heterogeneous OA paradigms. By employing a two-stage pipeline combining voting-based fusion strategies with post-fusion selection policies, the framework reconciles conflicting correspondences generated by diverse aligner algorithms. We systematically evaluated standalone and ensemble aligners across eight benchmark tasks from five OAEI tracks. Our experimental findings demonstrate that voting-based ensemble fusion consistently improves the precision--recall balance, frequently outperforming individual constituent baselines across varying domain complexities.

\begin{acknowledgments}
% This work is jointly supported by the \href{https://scinext-project.github.io/}{SCINEXT project} (BMFTR, German Federal Ministry of Research, Technology and Space, Grant ID: 01lS22070), the KISSKI AI Service Center (BMFTR, Grant ID: 01IS22093C), and the \href{https://www.nfdi4datascience.de/}{NFDI4DataScience initiative} (DFG, German Research Foundation, Grant ID: 460234259).
This work is supported by the  \href{https://www.nfdi4datascience.de/}{NFDI4DataScience initiative} (DFG, German Research Foundation, Grant ID: 460234259).

\end{acknowledgments}

%% The declaration on generative AI comes in effect
%% in Janary 2025. See also
%% https://ceur-ws.org/GenAI/Policy.html

\section*{Declaration on Generative AI}
In preparing this manuscript, generative AI tools, specifically ChatGPT, were used solely for grammar checking, spelling checks, and readability of some sentences.  All suggested changes were carefully reviewed and adapted by the authors to ensure accuracy and appropriateness. The scientific content, research design, analysis, and conclusions were developed and verified exclusively by the authors without AI involvement. The use of ChatGPT was limited to enhancing the presentation of the work.

Generative AI (GenAI) tools were used solely to assist with code
development for conducting the experiments reported in this paper
\bibliography{sample-ceur}

\begin{thebibliography}{41}
\expandafter\ifx\csname natexlab\endcsname\relax\def\natexlab#1{#1}\fi
\providecommand{\url}[1]{\texttt{#1}}
\providecommand{\href}[2]{#2}
\providecommand{\path}[1]{#1}
\providecommand{\DOIprefix}{doi:}
\providecommand{\ArXivprefix}{arXiv:}
\providecommand{\URLprefix}{URL: }
\providecommand{\Pubmedprefix}{pmid:}
\providecommand{\doi}[1]{\href{http://dx.doi.org/#1}{\path{#1}}}
\providecommand{\Pubmed}[1]{\href{pmid:#1}{\path{#1}}}
\providecommand{\bibinfo}[2]{#2}
\ifx\xfnm\relax \def\xfnm[#1]{\unskip,\space#1}\fi
%Type = Article
\bibitem[{Rahm and Bernstein(2001)}]{rahm2001survey}
\bibinfo{author}{E.~Rahm}, \bibinfo{author}{P.~A. Bernstein},
\newblock \bibinfo{title}{A survey of approaches to automatic schema matching},
\newblock \bibinfo{journal}{the VLDB Journal} \bibinfo{volume}{10} (\bibinfo{year}{2001}) \bibinfo{pages}{334--350}.
%Type = Inproceedings
\bibitem[{Milo et~al.(1998)Milo, Zohar et~al.}]{milo1998using}
\bibinfo{author}{T.~Milo}, \bibinfo{author}{S.~Zohar}, et~al.,
\newblock \bibinfo{title}{Using schema matching to simplify heterogeneous data translation},
\newblock in: \bibinfo{booktitle}{Vldb}, volume~\bibinfo{volume}{98}, \bibinfo{year}{1998}, pp. \bibinfo{pages}{24--27}.
%Type = Inproceedings
\bibitem[{Melnik et~al.(2002)Melnik, Garcia-Molina, and Rahm}]{melnik2002similarity}
\bibinfo{author}{S.~Melnik}, \bibinfo{author}{H.~Garcia-Molina}, \bibinfo{author}{E.~Rahm},
\newblock \bibinfo{title}{Similarity flooding: A versatile graph matching algorithm and its application to schema matching},
\newblock in: \bibinfo{booktitle}{Proceedings 18th international conference on data engineering}, \bibinfo{organization}{IEEE}, \bibinfo{year}{2002}, pp. \bibinfo{pages}{117--128}.
%Type = Inproceedings
\bibitem[{Noy and Musen(2000)}]{PROMPT2000}
\bibinfo{author}{N.~F. Noy}, \bibinfo{author}{M.~A. Musen},
\newblock \bibinfo{title}{Prompt: Algorithm and tool for automated ontology merging and alignment},
\newblock in: \bibinfo{booktitle}{Proceedings of the Seventeenth National Conference on Artificial Intelligence and Twelfth Conference on Innovative Applications of Artificial Intelligence}, \bibinfo{publisher}{AAAI Press}, \bibinfo{year}{2000}, p. \bibinfo{pages}{450–455}.
%Type = Inproceedings
\bibitem[{Noy and Musen(2001)}]{noy2001anchor}
\bibinfo{author}{N.~F. Noy}, \bibinfo{author}{M.~A. Musen},
\newblock \bibinfo{title}{Anchor-prompt: Using non-local context for semantic matching.},
\newblock in: \bibinfo{booktitle}{Ois@ ijcai}, \bibinfo{year}{2001}.
%Type = Inproceedings
\bibitem[{Doan et~al.(2002)Doan, Madhavan, Domingos, and Halevy}]{doan2002learning}
\bibinfo{author}{A.~Doan}, \bibinfo{author}{J.~Madhavan}, \bibinfo{author}{P.~Domingos}, \bibinfo{author}{A.~Halevy},
\newblock \bibinfo{title}{Learning to map between ontologies on the semantic web},
\newblock in: \bibinfo{booktitle}{Proceedings of the 11th international conference on World Wide Web}, \bibinfo{year}{2002}, pp. \bibinfo{pages}{662--673}.
%Type = Article
\bibitem[{Kalfoglou and Schorlemmer(2003)}]{kalfoglou2003ontology}
\bibinfo{author}{Y.~Kalfoglou}, \bibinfo{author}{M.~Schorlemmer},
\newblock \bibinfo{title}{Ontology mapping: the state of the art},
\newblock \bibinfo{journal}{The knowledge engineering review} \bibinfo{volume}{18} (\bibinfo{year}{2003}) \bibinfo{pages}{1--31}.
%Type = Book
\bibitem[{Euzenat and Shvaiko(2013)}]{euzenat2013ontology}
\bibinfo{author}{J.~Euzenat}, \bibinfo{author}{P.~Shvaiko}, \bibinfo{title}{Ontology Matching}, \bibinfo{edition}{2} ed., \bibinfo{publisher}{Springer}, \bibinfo{address}{Berlin, Heidelberg}, \bibinfo{year}{2013}. \DOIprefix\doi{10.1007/978-3-642-38721-0}.
%Type = Inproceedings
\bibitem[{Faria et~al.(2013)Faria, Pesquita, Santos, Palmonari, Cruz, and Couto}]{faria2013agreementmakerlight}
\bibinfo{author}{D.~Faria}, \bibinfo{author}{C.~Pesquita}, \bibinfo{author}{E.~Santos}, \bibinfo{author}{M.~Palmonari}, \bibinfo{author}{I.~F. Cruz}, \bibinfo{author}{F.~M. Couto},
\newblock \bibinfo{title}{The agreementmakerlight ontology matching system},
\newblock in: \bibinfo{booktitle}{OTM Confederated International Conferences" On the Move to Meaningful Internet Systems"}, \bibinfo{organization}{Springer}, \bibinfo{year}{2013}, pp. \bibinfo{pages}{527--541}.
%Type = Inproceedings
\bibitem[{Jim{\'e}nez-Ruiz and Cuenca~Grau(2011)}]{jimenez2011logmap}
\bibinfo{author}{E.~Jim{\'e}nez-Ruiz}, \bibinfo{author}{B.~Cuenca~Grau},
\newblock \bibinfo{title}{Logmap: Logic-based and scalable ontology matching},
\newblock in: \bibinfo{booktitle}{International Semantic Web Conference}, \bibinfo{organization}{Springer}, \bibinfo{year}{2011}, pp. \bibinfo{pages}{273--288}.
%Type = Article
\bibitem[{Li et~al.(2008)Li, Tang, Li, and Luo}]{li2008rimom}
\bibinfo{author}{J.~Li}, \bibinfo{author}{J.~Tang}, \bibinfo{author}{Y.~Li}, \bibinfo{author}{Q.~Luo},
\newblock \bibinfo{title}{Rimom: A dynamic multistrategy ontology alignment framework},
\newblock \bibinfo{journal}{IEEE Transactions on Knowledge and data Engineering} \bibinfo{volume}{21} (\bibinfo{year}{2008}) \bibinfo{pages}{1218--1232}.
%Type = Inproceedings
\bibitem[{Isaac et~al.(2007)Isaac, Van Der~Meij, Schlobach, and Wang}]{isaac2007instance}
\bibinfo{author}{A.~Isaac}, \bibinfo{author}{L.~Van Der~Meij}, \bibinfo{author}{S.~Schlobach}, \bibinfo{author}{S.~Wang},
\newblock \bibinfo{title}{An empirical study of instance-based ontology matching},
\newblock in: \bibinfo{booktitle}{International Semantic Web Conference}, \bibinfo{organization}{Springer}, \bibinfo{year}{2007}, pp. \bibinfo{pages}{253--266}.
%Type = Inproceedings
\bibitem[{Kolyvakis et~al.(2018)Kolyvakis, Kalousis, and Kiritsis}]{kolyvakis2018deepalignment}
\bibinfo{author}{P.~Kolyvakis}, \bibinfo{author}{A.~Kalousis}, \bibinfo{author}{D.~Kiritsis},
\newblock \bibinfo{title}{Deepalignment: Unsupervised ontology matching with refined word vectors},
\newblock in: \bibinfo{booktitle}{Proceedings of the 2018 Conference of the North American Chapter of the Association for Computational Linguistics: Human Language Technologies, Volume 1 (Long Papers)}, \bibinfo{year}{2018}, pp. \bibinfo{pages}{787--798}.
%Type = Article
\bibitem[{Chen et~al.(2021)Chen, Hu, Jimenez-Ruiz, Holter, Antonyrajah, and Horrocks}]{chen2021owl2vec}
\bibinfo{author}{J.~Chen}, \bibinfo{author}{P.~Hu}, \bibinfo{author}{E.~Jimenez-Ruiz}, \bibinfo{author}{O.~M. Holter}, \bibinfo{author}{D.~Antonyrajah}, \bibinfo{author}{I.~Horrocks},
\newblock \bibinfo{title}{Owl2vec*: embedding of owl ontologies},
\newblock \bibinfo{journal}{Machine Learning} \bibinfo{volume}{110} (\bibinfo{year}{2021}) \bibinfo{pages}{1813--1845}.
%Type = Inproceedings
\bibitem[{Iyer et~al.(2020)Iyer, Agarwal, and Kumar}]{iyer2020veealign}
\bibinfo{author}{V.~Iyer}, \bibinfo{author}{A.~Agarwal}, \bibinfo{author}{H.~Kumar},
\newblock \bibinfo{title}{Veealign: a supervised deep learning approach to ontology alignment.},
\newblock in: \bibinfo{booktitle}{OM@ ISWC}, \bibinfo{year}{2020}, pp. \bibinfo{pages}{216--224}.
%Type = Inproceedings
\bibitem[{He et~al.(2022)He, Chen, Antonyrajah, and Horrocks}]{he2022bertmap}
\bibinfo{author}{Y.~He}, \bibinfo{author}{J.~Chen}, \bibinfo{author}{D.~Antonyrajah}, \bibinfo{author}{I.~Horrocks},
\newblock \bibinfo{title}{Bertmap: a bert-based ontology alignment system},
\newblock in: \bibinfo{booktitle}{Proceedings of the AAAI Conference on Artificial Intelligence}, volume~\bibinfo{volume}{36}, \bibinfo{year}{2022}, pp. \bibinfo{pages}{5684--5691}.
%Type = Inproceedings
\bibitem[{Hertling and Paulheim(2023)}]{hertling2023olala}
\bibinfo{author}{S.~Hertling}, \bibinfo{author}{H.~Paulheim},
\newblock \bibinfo{title}{Olala: Ontology matching with large language models},
\newblock in: \bibinfo{booktitle}{Proceedings of the 12th knowledge capture conference 2023}, \bibinfo{year}{2023}, pp. \bibinfo{pages}{131--139}.
%Type = Article
\bibitem[{Qiang et~al.(2023)Qiang, Wang, and Taylor}]{qiang2023agent}
\bibinfo{author}{Z.~Qiang}, \bibinfo{author}{W.~Wang}, \bibinfo{author}{K.~Taylor},
\newblock \bibinfo{title}{Agent-om: Leveraging llm agents for ontology matching},
\newblock \bibinfo{journal}{arXiv preprint arXiv:2312.00326}  (\bibinfo{year}{2023}).
%Type = Inproceedings
\bibitem[{Babaei~Giglou et~al.(2024)Babaei~Giglou, D’Souza, Engel, and Auer}]{babaei2024llms4om}
\bibinfo{author}{H.~Babaei~Giglou}, \bibinfo{author}{J.~D’Souza}, \bibinfo{author}{F.~Engel}, \bibinfo{author}{S.~Auer},
\newblock \bibinfo{title}{Llms4om: Matching ontologies with large language models},
\newblock in: \bibinfo{booktitle}{European Semantic Web Conference}, \bibinfo{organization}{Springer}, \bibinfo{year}{2024}, pp. \bibinfo{pages}{25--35}.
%Type = Inproceedings
\bibitem[{Babaei~Giglou et~al.(2025)Babaei~Giglou, D’Souza, Karras, and Auer}]{babaei2025ontoaligner}
\bibinfo{author}{H.~Babaei~Giglou}, \bibinfo{author}{J.~D’Souza}, \bibinfo{author}{O.~Karras}, \bibinfo{author}{S.~Auer},
\newblock \bibinfo{title}{Ontoaligner: A comprehensive modular and robust python toolkit for ontology alignment},
\newblock in: \bibinfo{booktitle}{European Semantic Web Conference}, \bibinfo{year}{2025}, pp. \bibinfo{pages}{174--191}.
%Type = Misc
\bibitem[{Babaei~Giglou et~al.(2026)Babaei~Giglou, D'Souza, Karras, and Auer}]{babaei_giglou_2026_21206300}
\bibinfo{author}{H.~Babaei~Giglou}, \bibinfo{author}{J.~D'Souza}, \bibinfo{author}{O.~Karras}, \bibinfo{author}{S.~Auer}, \bibinfo{title}{Ontoaligner: A comprehensive modular and robust python toolkit for ontology alignment}, \bibinfo{year}{2026}. \URLprefix \url{https://doi.org/10.5281/zenodo.21206300}. \DOIprefix\doi{10.5281/zenodo.21206300}.
%Type = Article
\bibitem[{Giglou et~al.(2025)Giglou, D'Souza, Auer, and Sanaei}]{giglou2025ontoaligner}
\bibinfo{author}{H.~B. Giglou}, \bibinfo{author}{J.~D'Souza}, \bibinfo{author}{S.~Auer}, \bibinfo{author}{M.~Sanaei},
\newblock \bibinfo{title}{Ontoaligner meets knowledge graph embedding aligners},
\newblock \bibinfo{journal}{arXiv preprint arXiv:2509.26417}  (\bibinfo{year}{2025}).
%Type = Inproceedings
\bibitem[{Khan et~al.(2023)Khan, Saqib, Khattak, Ali, and Lee}]{khan2023ontology}
\bibinfo{author}{H.~Khan}, \bibinfo{author}{M.~Saqib}, \bibinfo{author}{H.~A. Khattak}, \bibinfo{author}{S.~I. Ali}, \bibinfo{author}{S.~Lee},
\newblock \bibinfo{title}{Ontology alignment for accurate ontology matching: A survey},
\newblock in: \bibinfo{booktitle}{International Conference on Smart Homes and Health Telematics}, \bibinfo{organization}{Springer}, \bibinfo{year}{2023}, pp. \bibinfo{pages}{338--349}.
%Type = Inproceedings
\bibitem[{Eckert et~al.(2009)Eckert, Meilicke, and Stuckenschmidt}]{5Eckert}
\bibinfo{author}{K.~Eckert}, \bibinfo{author}{C.~Meilicke}, \bibinfo{author}{H.~Stuckenschmidt},
\newblock \bibinfo{title}{Improving ontology matching using meta-level learning},
\newblock in: \bibinfo{editor}{L.~Aroyo}, \bibinfo{editor}{P.~Traverso}, \bibinfo{editor}{F.~Ciravegna}, \bibinfo{editor}{P.~Cimiano}, \bibinfo{editor}{T.~Heath}, \bibinfo{editor}{E.~Hyv{\"o}nen}, \bibinfo{editor}{R.~Mizoguchi}, \bibinfo{editor}{E.~Oren}, \bibinfo{editor}{M.~Sabou}, \bibinfo{editor}{E.~Simperl} (Eds.), \bibinfo{booktitle}{The Semantic Web: Research and Applications}, \bibinfo{publisher}{Springer Berlin Heidelberg}, \bibinfo{address}{Berlin, Heidelberg}, \bibinfo{year}{2009}, pp. \bibinfo{pages}{158--172}.
%Type = Inproceedings
\bibitem[{Nkisi-Orji et~al.(2019)Nkisi-Orji, Wiratunga, Massie, Hui, and Heaven}]{6Nkuisi}
\bibinfo{author}{I.~Nkisi-Orji}, \bibinfo{author}{N.~Wiratunga}, \bibinfo{author}{S.~Massie}, \bibinfo{author}{K.-Y. Hui}, \bibinfo{author}{R.~Heaven},
\newblock \bibinfo{title}{Ontology alignment based on word embedding and random forest classification},
\newblock in: \bibinfo{editor}{M.~Berlingerio}, \bibinfo{editor}{F.~Bonchi}, \bibinfo{editor}{T.~G{\"a}rtner}, \bibinfo{editor}{N.~Hurley}, \bibinfo{editor}{G.~Ifrim} (Eds.), \bibinfo{booktitle}{Machine Learning and Knowledge Discovery in Databases}, \bibinfo{publisher}{Springer International Publishing}, \bibinfo{address}{Cham}, \bibinfo{year}{2019}, pp. \bibinfo{pages}{557--572}.
%Type = Inproceedings
\bibitem[{Ding et~al.(2025)Ding, Wang, Liu, Han, and Wang}]{1Ding}
\bibinfo{author}{J.~Ding}, \bibinfo{author}{C.~Wang}, \bibinfo{author}{J.~Liu}, \bibinfo{author}{S.~Han}, \bibinfo{author}{T.~Wang},
\newblock \bibinfo{title}{Rome: A robust ontology matching method based on ensemble learning},
\newblock in: \bibinfo{booktitle}{2025 IEEE 11th Conference on Big Data Security on Cloud (BigDataSecurity)}, \bibinfo{year}{2025}, pp. \bibinfo{pages}{121--125}. \DOIprefix\doi{10.1109/BigDataSecurity66063.2025.00022}.
%Type = Article
\bibitem[{Xue et~al.(2026)Xue, Chun-Wei~Lin, and Jiang}]{2Xue}
\bibinfo{author}{X.~Xue}, \bibinfo{author}{J.~Chun-Wei~Lin}, \bibinfo{author}{Z.~Jiang},
\newblock \bibinfo{title}{Collaborative ontology matching with dual population genetic programming and active meta-learning},
\newblock \bibinfo{journal}{IEEE Transactions on Evolutionary Computation} \bibinfo{volume}{30} (\bibinfo{year}{2026}) \bibinfo{pages}{1024--1038}. \DOIprefix\doi{10.1109/TEVC.2025.3569336}.
%Type = Misc
\bibitem[{Cheng et~al.(2024)Cheng, Fürst, Jacobs, and Garrido-Hidalgo}]{7Cheng}
\bibinfo{author}{B.~Cheng}, \bibinfo{author}{J.~Fürst}, \bibinfo{author}{T.~Jacobs}, \bibinfo{author}{C.~Garrido-Hidalgo}, \bibinfo{title}{Interactive ontology matching with cost-efficient learning}, \bibinfo{year}{2024}. \URLprefix \url{https://arxiv.org/abs/2404.07663}. \href{http://arxiv.org/abs/2404.07663}{{\tt arXiv:2404.07663}}.
%Type = Inproceedings
\bibitem[{Dearing and Goan(2017)}]{4dearing2017exploring}
\bibinfo{author}{D.~Dearing}, \bibinfo{author}{T.~Goan},
\newblock \bibinfo{title}{Exploring the synergies between biocuration and ontology alignment automation.},
\newblock in: \bibinfo{booktitle}{OM@ ISWC}, \bibinfo{year}{2017}, pp. \bibinfo{pages}{25--36}.
%Type = Article
\bibitem[{Gharpure et~al.(2024)}]{gharpure2024hybrid}
\bibinfo{author}{P.~Gharpure}, et~al.,
\newblock \bibinfo{title}{Hybrid approach to instance matching}  (\bibinfo{year}{2024}).
%Type = Inproceedings
\bibitem[{Hertling and Paulheim(2023)}]{hertling2023transformer}
\bibinfo{author}{S.~Hertling}, \bibinfo{author}{H.~Paulheim},
\newblock \bibinfo{title}{Transformer based semantic relation typing for knowledge graph integration},
\newblock in: \bibinfo{booktitle}{European Semantic Web Conference}, \bibinfo{organization}{Springer}, \bibinfo{year}{2023}, pp. \bibinfo{pages}{105--121}.
%Type = Article
\bibitem[{Arnold and Rahm(2014)}]{arnold2014enriching}
\bibinfo{author}{P.~Arnold}, \bibinfo{author}{E.~Rahm},
\newblock \bibinfo{title}{Enriching ontology mappings with semantic relations},
\newblock \bibinfo{journal}{Data \& Knowledge Engineering} \bibinfo{volume}{93} (\bibinfo{year}{2014}) \bibinfo{pages}{1--18}.
%Type = Inproceedings
\bibitem[{Blomqvist et~al.(2023)Blomqvist, Li, Keskis{\"a}rkk{\"a}, Lindecrantz, Pour, Li, and Lambrix}]{blomqvist2023cross}
\bibinfo{author}{E.~Blomqvist}, \bibinfo{author}{H.~Li}, \bibinfo{author}{R.~Keskis{\"a}rkk{\"a}}, \bibinfo{author}{M.~Lindecrantz}, \bibinfo{author}{M.~A.~N. Pour}, \bibinfo{author}{Y.~Li}, \bibinfo{author}{P.~Lambrix},
\newblock \bibinfo{title}{Cross-domain modelling-a network of core ontologies for the circular economy.},
\newblock in: \bibinfo{booktitle}{WOP@ ISWC}, \bibinfo{year}{2023}, pp. \bibinfo{pages}{1--12}.
%Type = Article
\bibitem[{Dragisic et~al.(2017)Dragisic, Ivanova, Li, and Lambrix}]{anatomy}
\bibinfo{author}{Z.~Dragisic}, \bibinfo{author}{V.~Ivanova}, \bibinfo{author}{H.~Li}, \bibinfo{author}{P.~Lambrix},
\newblock \bibinfo{title}{Experiences from the anatomy track in the ontology alignment evaluation initiative},
\newblock \bibinfo{journal}{J Biomed Semant} \bibinfo{volume}{8} (\bibinfo{year}{2017}) \bibinfo{pages}{56}. \URLprefix \url{https://doi.org/10.1186/s13326-017-0166-5}. \DOIprefix\doi{10.1186/s13326-017-0166-5}.
%Type = Misc
\bibitem[{Nas and Huschka(2023)}]{mse}
\bibinfo{author}{E.~Nas}, \bibinfo{author}{M.~Huschka}, \bibinfo{title}{{MSE Benchmark}}, \bibinfo{howpublished}{\url{https://github.com/EngyNasr/MSE-Benchmark}}, \bibinfo{year}{2023}.
%Type = Article
\bibitem[{Karam et~al.(2020)Karam, Khiat, Algergawy, Sattler, Weiland, and Schmidt}]{biodiversity}
\bibinfo{author}{N.~Karam}, \bibinfo{author}{A.~Khiat}, \bibinfo{author}{A.~Algergawy}, \bibinfo{author}{M.~Sattler}, \bibinfo{author}{C.~Weiland}, \bibinfo{author}{M.~Schmidt},
\newblock \bibinfo{title}{Matching biodiversity and ecology ontologies: challenges and evaluation results},
\newblock \bibinfo{journal}{The Knowledge Engineering Review} \bibinfo{volume}{35} (\bibinfo{year}{2020}) \bibinfo{pages}{e9}. \URLprefix \url{https://doi.org/10.1017/S0269888920000132}. \DOIprefix\doi{10.1017/S0269888920000132}.
%Type = Misc
\bibitem[{{Qwen Team}(2026)}]{qwen3.5}
\bibinfo{author}{{Qwen Team}}, \bibinfo{title}{{Qwen3.5}: Towards native multimodal agents}, \bibinfo{year}{2026}. \URLprefix \url{https://qwen.ai/blog?id=qwen3.5}.
%Type = Article
\bibitem[{Zhang et~al.(2025)Zhang, Li, Long, Zhang, Lin, Yang, Xie, Yang, Liu, Lin, Huang, and Zhou}]{qwen3embedding}
\bibinfo{author}{Y.~Zhang}, \bibinfo{author}{M.~Li}, \bibinfo{author}{D.~Long}, \bibinfo{author}{X.~Zhang}, \bibinfo{author}{H.~Lin}, \bibinfo{author}{B.~Yang}, \bibinfo{author}{P.~Xie}, \bibinfo{author}{A.~Yang}, \bibinfo{author}{D.~Liu}, \bibinfo{author}{J.~Lin}, \bibinfo{author}{F.~Huang}, \bibinfo{author}{J.~Zhou},
\newblock \bibinfo{title}{Qwen3 embedding: Advancing text embedding and reranking through foundation models},
\newblock \bibinfo{journal}{arXiv preprint arXiv:2506.05176}  (\bibinfo{year}{2025}).
%Type = Article
\bibitem[{Schechter~Vera et~al.(2025)Schechter~Vera, Dua, Zhang, Salz, Mullins, Raghuram~Panyam, Smoot, Naim, Zou, Chen, Cer, Lisak, Choi, Gonzalez, Sanseviero, Cameron, Ballantyne, Black, Chen, Wang, Li, Martins, Lee, Sherwood, Ji, Wu, Zheng, Singh, Sharma, Sreepat, Jain, Elarabawy, Co, Doumanoglou, Samari, Hora, Potetz, Kim, Alfonseca, Moiseev, Han, Palma~Gomez, Hernández~Ábrego, Zhang, Hui, Han, Gill, Chen, Chen, Shanbhogue, Boratko, Suganthan, Duddu, Mariserla, Ariafar, Zhang, Zhang, Baumgartner, Goenka, Qiu, Dabral, Walker, Rao, Khawaja, Zhou, Ren, Xia, Chen, Chen, Dong, Ding, Visin, Liu, Zhang, Kenealy, Casbon, Kumar, Mesnard, Gleicher, Brick, Lacombe, Roberts, Sung, Hoffmann, Warkentin, Joulin, Duerig, and Seyedhosseini}]{embedding_gemma_2025}
\bibinfo{author}{H.~Schechter~Vera}, \bibinfo{author}{S.~Dua}, \bibinfo{author}{B.~Zhang}, \bibinfo{author}{D.~Salz}, \bibinfo{author}{R.~Mullins}, \bibinfo{author}{S.~Raghuram~Panyam}, \bibinfo{author}{S.~Smoot}, \bibinfo{author}{I.~Naim}, \bibinfo{author}{J.~Zou}, \bibinfo{author}{F.~Chen}, \bibinfo{author}{D.~Cer}, \bibinfo{author}{A.~Lisak}, \bibinfo{author}{M.~Choi}, \bibinfo{author}{L.~Gonzalez}, \bibinfo{author}{O.~Sanseviero}, \bibinfo{author}{G.~Cameron}, \bibinfo{author}{I.~Ballantyne}, \bibinfo{author}{K.~Black}, \bibinfo{author}{K.~Chen}, \bibinfo{author}{W.~Wang}, \bibinfo{author}{Z.~Li}, \bibinfo{author}{G.~Martins}, \bibinfo{author}{J.~Lee}, \bibinfo{author}{M.~Sherwood}, \bibinfo{author}{J.~Ji}, \bibinfo{author}{R.~Wu}, \bibinfo{author}{J.~Zheng}, \bibinfo{author}{J.~Singh}, \bibinfo{author}{A.~Sharma}, \bibinfo{author}{D.~Sreepat}, \bibinfo{author}{A.~Jain}, \bibinfo{author}{A.~Elarabawy}, \bibinfo{author}{A.~Co}, \bibinfo{author}{A.~Doumanoglou}, \bibinfo{author}{B.~Samari},
  \bibinfo{author}{B.~Hora}, \bibinfo{author}{B.~Potetz}, \bibinfo{author}{D.~Kim}, \bibinfo{author}{E.~Alfonseca}, \bibinfo{author}{F.~Moiseev}, \bibinfo{author}{F.~Han}, \bibinfo{author}{F.~Palma~Gomez}, \bibinfo{author}{G.~Hernández~Ábrego}, \bibinfo{author}{H.~Zhang}, \bibinfo{author}{H.~Hui}, \bibinfo{author}{J.~Han}, \bibinfo{author}{K.~Gill}, \bibinfo{author}{K.~Chen}, \bibinfo{author}{K.~Chen}, \bibinfo{author}{M.~Shanbhogue}, \bibinfo{author}{M.~Boratko}, \bibinfo{author}{P.~Suganthan}, \bibinfo{author}{S.~M.~K. Duddu}, \bibinfo{author}{S.~Mariserla}, \bibinfo{author}{S.~Ariafar}, \bibinfo{author}{S.~Zhang}, \bibinfo{author}{S.~Zhang}, \bibinfo{author}{S.~Baumgartner}, \bibinfo{author}{S.~Goenka}, \bibinfo{author}{S.~Qiu}, \bibinfo{author}{T.~Dabral}, \bibinfo{author}{T.~Walker}, \bibinfo{author}{V.~Rao}, \bibinfo{author}{W.~Khawaja}, \bibinfo{author}{W.~Zhou}, \bibinfo{author}{X.~Ren}, \bibinfo{author}{Y.~Xia}, \bibinfo{author}{Y.~Chen}, \bibinfo{author}{Y.-T. Chen}, \bibinfo{author}{Z.~Dong},
  \bibinfo{author}{Z.~Ding}, \bibinfo{author}{F.~Visin}, \bibinfo{author}{G.~Liu}, \bibinfo{author}{J.~Zhang}, \bibinfo{author}{K.~Kenealy}, \bibinfo{author}{M.~Casbon}, \bibinfo{author}{R.~Kumar}, \bibinfo{author}{T.~Mesnard}, \bibinfo{author}{Z.~Gleicher}, \bibinfo{author}{C.~Brick}, \bibinfo{author}{O.~Lacombe}, \bibinfo{author}{A.~Roberts}, \bibinfo{author}{Y.~Sung}, \bibinfo{author}{R.~Hoffmann}, \bibinfo{author}{T.~Warkentin}, \bibinfo{author}{A.~Joulin}, \bibinfo{author}{T.~Duerig}, \bibinfo{author}{M.~Seyedhosseini},
\newblock \bibinfo{title}{Embeddinggemma: Powerful and lightweight text representations}  (\bibinfo{year}{2025}). \URLprefix \url{https://arxiv.org/abs/2509.20354}.
%Type = Inproceedings
\bibitem[{Liu et~al.(2024)Liu, Grode, and Hansen}]{liu2024mdmapper}
\bibinfo{author}{X.~Liu}, \bibinfo{author}{J.~Grode}, \bibinfo{author}{M.~R. Hansen},
\newblock \bibinfo{title}{Mdmapper: A framework for aligning master data models using ontology matching techniques},
\newblock in: \bibinfo{booktitle}{The 19th International Workshop on Ontology Matching}, \bibinfo{organization}{CEUR-WS}, \bibinfo{year}{2024}, pp. \bibinfo{pages}{30--42}.
%Type = Article
\bibitem[{Cotovio et~al.(2024)Cotovio, Ferraz, Faria, Balbi, Silva, and Pesquita}]{cotovio2024matcha}
\bibinfo{author}{P.~G. Cotovio}, \bibinfo{author}{L.~Ferraz}, \bibinfo{author}{D.~Faria}, \bibinfo{author}{L.~Balbi}, \bibinfo{author}{M.~C. Silva}, \bibinfo{author}{C.~Pesquita},
\newblock \bibinfo{title}{Matcha-dl a tool for supervised ontology alignment},
\newblock \bibinfo{journal}{preprint}  (\bibinfo{year}{2024}).

\end{thebibliography}

\end{document}